\documentclass{article}
\usepackage{microtype}
\usepackage{multirow}
\usepackage{graphicx}
\usepackage{subfigure}
\usepackage{booktabs} 

\usepackage{hyperref}

\usepackage[capitalise]{cleveref}
\usepackage[section]{placeins}
\usepackage{flafter}

\usepackage[accepted]{mlsys2023_arxiv}

\mlsystitlerunning{Privacy in Multimodal Federated Human Activity Recognition}

\begin{document}

\twocolumn[
    \mlsystitle{Privacy in Multimodal Federated Human Activity Recognition}

    \chead{PREPRINT: Accepted at The 3rd On-Device Intelligence Workshop@MLSys 23}

    \fancypagestyle{firstpagestyle}{%
        \fancyhf{}
        \chead{PREPRINT: Accepted at The 3rd On-Device Intelligence Workshop@MLSys 23}
        \renewcommand{\headrulewidth}{0.0pt}%

    }




    \begin{mlsysauthorlist}
        \mlsysauthor{Alex Iacob}{cam}
        \mlsysauthor{Pedro P. B. Gusmão}{cam}
        \mlsysauthor{Nicholas D. Lane}{cam}
        \mlsysauthor{Armand K. Koupai}{brist}
        \mlsysauthor{Mohammud J. Bocus}{brist}
        \mlsysauthor{Raúl Santos-Rodríguez}{brist}
        \mlsysauthor{Robert J. Piechocki}{brist}
        \mlsysauthor{Ryan McConville}{brist}
    \end{mlsysauthorlist}

    \mlsysaffiliation{cam}{Department of Computer Science and Technology, University of Cambridge, United Kingdom}
    \mlsysaffiliation{brist}{School of Computer Science, Electrical and Electronic Engineering, and Engineering Maths,
        University of Bristol, United Kingdom}

    \mlsyscorrespondingauthor{Alex Iacob}{aai30@cam.ac.uk}
    \mlsyscorrespondingauthor{Pedro Porto Buarque de Gusmao}{pp524@cam.ac.uk}
    \mlsyscorrespondingauthor{Nicholas Donald Lane}{ndl32@cam.ac.uk}
    \mlsyscorrespondingauthor{Armand K. Koupai}{uw20504@bristol.ac.uk}
    \mlsyscorrespondingauthor{Mohammud J. Bocus}{junaid.bocus@bristol.ac.uk}
    \mlsyscorrespondingauthor{Raúl Santos-Rodríguez}{enrsr@bristol.ac.uk}
    \mlsyscorrespondingauthor{Robert J. Piechocki}{eerjp@bristol.ac.uk}
    \mlsyscorrespondingauthor{Ryan McConville}{ryan.mcconville@bristol.ac.uk}

    \mlsyskeywords{Machine Learning, MLSys, Federated Learning, Human Activity Recognition, Privacy}

    \vskip 0.3in

    \begin{abstract}
        Human Activity Recognition (HAR) training data is often privacy-sensitive or held by non-cooperative entities. Federated Learning (FL) addresses such concerns by training ML models on edge clients. This work studies the impact of privacy in federated HAR at a user, environment, and sensor level. We show that the performance of FL for HAR depends on the assumed privacy level of the FL system and primarily upon the colocation of data from different sensors. By avoiding data sharing and assuming privacy at the human or environment level, as prior works have done, the accuracy decreases by $5\mbox{-}7\%$. However, extending this to the modality level and strictly separating sensor data between multiple clients may decrease the accuracy by $19\mbox{-}42\%$. As this form of privacy is necessary for the ethical utilisation of passive sensing methods in HAR, we implement a system where clients mutually train both a general FL model and a group-level one per modality. Our evaluation shows that this method leads to only a $7\mbox{-}13\%$ decrease in accuracy, making it possible to build HAR systems with diverse hardware.
    \end{abstract}
    \thispagestyle{firstpagestyle}
]


\printAffiliationsAndNotice{}

\section{Introduction}
\label{Introduction}
Human Activity Recognition (HAR) involves classifying human actions~\citep{HARSurvey,RecentHARSurvey}, such as running or sitting, using data from personal devices like smartphones or environmental sensors. However, practical and legal considerations limit learning from HAR data. For example, using video cameras to simulate virtual bodily-worn movement sensors~\citep{IMUTube} may generate divergent features from Wi-Fi signals. Furthermore, privacy requirements impose data collection limitations. In this work, privacy requirements refer to constraints on collecting or centralising data at three levels:

\textbf{User(Human subject)-level Privacy} For gyroscope or accelerometer data from smartphones and wearables, end-users may be unwilling to share personal information.

\textbf{Environment-level Privacy} For locations such as hospitals and internment facilities, sensitive information must often remain private from third parties. This constraint may prove challenging as data used for HAR is susceptible to environmental characteristics. For example, a sensor may produce varying features based on object placement or room size.

\textbf{Modality-level Privacy} Data generated from different groups of sensors may be owned by competing entities.

Traditional Machine Learning approaches tackle feature heterogeneity by colocating data and training with Multi-task Learning techniques. However, the privacy constraints above make centralisation unfeasible on a large scale in HAR\@. Instead, they require a Federated Learning (FL) approach to keep data encapsulated in clients at the necessary privacy level during training. Our work brings the following contributions to Federated Human Activity Recognition:
\vspace{-0.24cm}
\begin{enumerate}
    \item First, we evaluate the performance of multiple models trained in a federated fashion on a multimodal dataset keeping data privately stored on clients at increasing privacy levels. Unlike other works, we investigate the additive effects of privacy up to the complete separation of each user, environment, and modality combination.
    \item Second, we show that privacy at the modality level results in the \emph{highest} accuracy cost, followed by the environmental level and then the user level. To mitigate this, we propose mutual learning of group-level models alongside the standard FL model to cover modalities that cannot be colocated in a single client. Our results indicate that this method can significantly reduce accuracy degradation from $19\mbox{-}42\%$ to just $7\mbox{-}13\%$.
\end{enumerate}

\section{Multimodality in Federated Human Activity Recognition}
\label{Challanges_Background}
Federated Learning, proposed by \citet{FedAvg}, trains ML models from distributed data on edge devices using efficient communication techniques for maintaining privacy. Although successful in training models from diverse users, such as keyboard prediction \citep{FLKeyboard}, and diverse hardware, such as medical applications \citep{FLmedicine}, data heterogeneity remains a significant challenge \citep[sec 3.1]{AdvancedAndOpenProblems}. Due to privacy constraints, approaches like Multi-task Learning and Continual Learning, which handle feature heterogeneity, are limited in a Federated Learning context. For instance, Elastic Weight Consolidation \citep{EWC} estimates parameter variance using past data, while Learning Without Forgetting \citep{LWF} stores network outputs from past tasks.

Previous work on Federated Human Activity Recognition has not fully explored the heterogeneity emerging from independent data collection systems. For instance, \citet{HARusingFL_2018} considers skewed label distributions and noise across smartphone users while \emph{colocating} gyroscope and accelerometer data. Similar partitioning schemes are investigated for feature extraction \citep{HARFLEnchancedFeature} and clustering methods \citep{ClusterFL}. Furthermore, such works may use artificially partitioned centralised datasets, as in \citet{HARsemisupervised}, or contain only one modality, as in some datasets collected by \citet{ClusterFL}. To create adaptable HAR systems that can accommodate new clients with different sensor types in the federation, investigating Federated HAR with modalities split across clients is necessary, given the shifting hardware landscape of HAR sensors.

\section{Federating Human Activity Recognition}
\label{Federating_Methods}
We construct multiple partitions of the OPERAnet dataset published by \citet{OPERAnet} to assess Federated Human Activity Recognition under privacy at user, environment, and modality privacy levels. The dataset contains five different sensors; however, \citet{OPERAnet} indicate that only Channel State Information (CSI) from a Network Card Interface (NIC) and Passive Wi-Fi Radar (PWR) should be used for HAR\@. The data were collected synchronously, with the multiple channels---three for CSI and two for PWR---of RF data. They cover eight hours of surveying six participants performing six activities spread across two rooms. Because room activity distribution is non-uniform, separating clients by environment also provides skewness at the label level.

We transform the time-series data into image data in keeping with the original HAR preprocessing applied by \citet{OPERAnet} and previous works~\citep{prev1,prev2,prev3}. We further increase the dataset's size and modality diversity by reusing the pipeline of \citet{SelfSupervisedVision}. Based on the underlying CSI and PWR data, \citet{SelfSupervisedVision} construct spectrograms of the CSI and PWR data. The complete image set contains five data views for each underlying CSI or PWR channel. We use the three most effective view types reported by \citet{SelfSupervisedVision}. Since different channels for CSI and PWR are physically colocated on the device, it is assumed that fusing images generated from separate channels would not be a violation of privacy at the sensor level. Consequently, the complete image types we shall refer to as modalities for the rest of this work contain concatenated images generated from each source channel. One such image type comes from CSI; two come from PWR.

The work of \citet{SelfSupervisedVision} offers two centralised baselines to compare against, a ResNet34~\citep{ResNet} model used for HAR and a Fusion Vision Transformer (FViT). While CNNs have been successfully applied to HAR by \citet{HARCnn,Inception-ResNet} and \citet{MultiscaleResnetHAR}, the novel FViT addresses the issue of multimodal HAR by adapting the Vision Transformer (ViT) architecture developed by \citet{VisionTransformer} to operate over \emph{fused} images. Crucially for our experiments, FViT has a parameter count invariant to the number of images combined, making the network capacity equivalent between fused and unfused modalities. In addition to the transformer and ResNet34, we use the smallest EfficientNetV2 constructed by \citet{EfficientNetV2} as the communication costs and compute concerns in FL make the smaller network a practical choice.

\subsection{Partitioning by Privacy Level}
The partitions we construct correspond to increasing privacy levels. For example, splitting by human subject implies that each client in that partition only contains data corresponding to one human participant and thus obeys \emph{Subject(User)-level} privacy. Likewise, the partition splitting by participant and room implies that each participant and room combination is treated as a separate client and offers both \emph{Subject(User)-level} and \emph{Environment-level} privacy. The most heterogeneous partition we create treats each participant, room, and modality combination as one client and offers the previous two levels of privacy together with \emph{Modality-level} privacy.

\begin{table}[t!]
    \centering
    \caption{The partitions in our experimental setup. They include the centralised baseline (\emph{Centralised}), those partitioned by human subject (\emph{Subj}), by subject and environment (\emph{Subj+Env}), or by subject, environment, and modality (\emph{Subj+Env+Mod}). A partition can contain fused (\emph{F}) or separated (\emph{S}) modalities.}
    \label{tab:ExperimentalSetup}
    \resizebox{\columnwidth}{!}{%
        \begin{tabular}{@{}lllll@{}}
            \toprule
            Partition            & Avg Samples       & \#Samples Subj 6 & \#Train Clients & \#Clients/R \\ \midrule
            Centralised (F)      & $1947.0 \pm 0.0$  & 463              & 1               & 1           \\
            Subj (F)             & $324.3 \pm 32.3$  & 463              & 6               & 2           \\
            Subj + Env (F)       & $194.6 \pm 194.6$ & 463              & 10              & 3           \\ \midrule
            Centralised (S)      & $5841.0 \pm 0.0$  & 1389             & 1               & 1           \\
            Subj (S)             & $973.0 \pm 973.0$ & 1389             & 6               & 2           \\
            Subj + Env (S)       & $583.8 \pm 583.8$ & 1389             & 10              & 3           \\
            Subj + Env + Mod (S) & $194.6 \pm 194.6$ & 1389             & 30              & 9           \\ \bottomrule
        \end{tabular}%
    }
\end{table}

To create a meaningful test set for Federated HAR, we use the data of the sixth client. Since OPERAnet has not been used for FL before, our evaluation compares the accuracy of FL partitioned by subject and environment to the State of The Art centralised baselines using colocated fused modalities. Following this initial investigation, we explore privacy interactions at a subject (user), environmental and modality level when the modalities are never fused and not necessarily colocated. The \emph{separated-modality} experiments are the ones we use to report findings, as they can cover all levels of privacy. \Cref{tab:ExperimentalSetup} presents the constructed partitions and their statistics. As we intended to use $30\%$ of total clients each round, we split the data of one participant into two clients based on their room when federating by subject. For the centralised baseline, we follow \citet{SelfSupervisedVision} and train for $100$ local epochs, while FL trains for $10$ rounds with $10$ local epochs. The optimiser parameters are kept constant and at parity with \citet{SelfSupervisedVision}---see \cref{Appendix:parameters}.

\subsection{Mutual Global and Group Model Learning}
To handle separating modalities across clients in a federated network, we propose a group FL structure utilising Deep Mutual Learning \citep{DeepMutualLearning}. Two models are trained on each client and distil knowledge into each other. One model is a globally federated model trained on all clients. The other is a group-level one trained only on clients with a specific modality. The server maintains one model per modality group, providing flexibility for integrating new sensors. We chose the FViT as the global federated model because of its resilience to high heterogeneity in previous experiments. Furthermore, we use the small EfficientNetV2 as the group-level model for future scalability. We present the performance of an ensemble of group-level models, each predicting the relevant modality. Hyperparameters were optimised via Bayesian search, resulting in global and group-level distillation weights of $0.33$ and $0.75$, respectively.

\section{Evaluation}\label{Evaluation}

\begin{table}[t]
    \centering
    \caption{Accuracy results (mean and standard deviation) for model and partition combinations on the test set of OPERAnet. Note the impact of partitioning by modality compared to the subject or environment and the smoother decline in the performance of FViT compared to the CNNs. The “Ensemble” uses the three group models to predict the data label belonging to their modality. The results of F1-Score, shown in \cref{tab:Appendix:F1-Score}, follow the same trend.}
    \label{tab:EvaluationAccuracy}
    \resizebox{\columnwidth}{!}{%
        \begin{tabular}{lllll}
            \hline
            Partition           & FViT                            & ResNet34                        & EffNetB0                        & Ensemble                        \\ \hline
            Centralised (Fused) & 0.90$\pm$0.01                   & \textbf{0.93}$\pm$\textbf{0.01} & 0.91$\pm$0.01                   & -                               \\
            Subj (Fused)        & 0.85$\pm$0.02                   & \textbf{0.90}$\pm$\textbf{0.02} & 0.88$\pm$0.01                   & -                               \\
            Subj+Env (Fused)    & 0.83$\pm$0.03                   & \textbf{0.84}$\pm$\textbf{0.07} & 0.79$\pm$0.04                   & -                               \\ \hline
            Centralised (S)     & 0.83$\pm$0.01                   & \textbf{0.89}$\pm$\textbf{0.02} & 0.87$\pm$0.01                   & -                               \\
            Subj (S)            & 0.81$\pm$0.01                   & 0.84$\pm$0.03                   & \textbf{0.85}$\pm$\textbf{0.01} & -                               \\
            Subj+Env (S)        & 0.78$\pm$0.02                   & \textbf{0.84}$\pm$\textbf{0.02} & 0.80$\pm$0.01                   & -                               \\
            Subj+Env+Mod (S)    & \textbf{0.64}$\pm$\textbf{0.04} & 0.47$\pm$0.03                   & 0.55$\pm$0.05                   & \textbf{0.76}$\pm$\textbf{0.02} \\ \hline
        \end{tabular}%
    }
\end{table}

\begin{figure*}[t]
    \centering
    \includegraphics[clip,width=\textwidth]{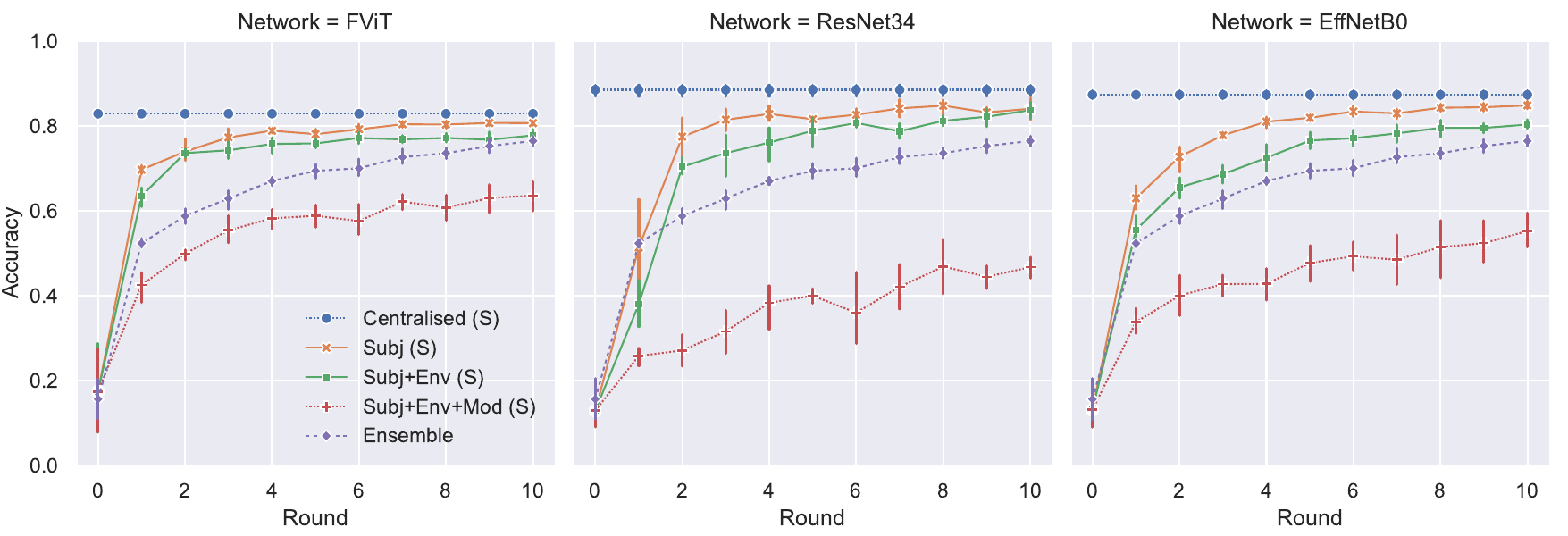}
    \caption[]{Model per-round accuracy on the sixth subject's dataset at all privacy levels. For low-heterogeneity, CNNs generally outperform FViT, but FViT gains a significant advantage when modality partitioning is introduced. Notably, FViT converges faster in the initial rounds and reaches higher accuracy with fewer data in the first two rounds. The “Ensemble” results reverse the effects of modality-level privacy and outperform all non-grouped federated models.}
    \label{fig:CorrImprovLoc}
\end{figure*}

Our evaluation reveals that Federated Human Activity Recognition is sensitive to sensor heterogeneity but partially resilient to subject characteristics and room structure. In \cref{tab:EvaluationAccuracy}, we present accuracy results of all partition and model combinations using fused (F) or separated (S) modalities. Fused modality experiments establish a baseline of comparison between federated and centralised training on OPERAnet. Separated modality experiments allow us to investigate more granular levels of privacy and will provide most of the notable figures. Accuracy convergence curves are available per model in \cref{fig:CorrImprovLoc}; however, they only show unfused modalities to emphasise results for modality-level privacy. In the appendix, convergence curves for fused modalities are presented in \cref{fig:AppendixFigFusedConvergence} and follow similar trends. All experiments used the Flower~\citep{Flower} FL framework.

\subsection{Subject(User)-level Privacy}

Experiments with user-level privacy showed that all three models achieved results within $3 \mbox{-} 5\%$ of the centralised baseline, regardless of modality fusion. As shown in \cref{fig:CorrImprovLoc}, this small gap to the centralised unfused baseline was consistently observed across rounds. In addition, a study by \citet{HARusingFL_2018} found a similar $4\mbox{-}6\%$ accuracy gap when treating each person as a separate partition, indicating that body characteristics and slight movement differences are not significant enough to generate highly divergent features. These findings suggest that FL is a practical solution for accessing extensive private data from end users. However, unfused modalities reduced accuracy for centralised and subject-level partitions nearly uniformly compared to fused ones---showcasing the benefits of centralisation.
\subsection{Environment-level Privacy}

A further slight-to-medium accuracy degradation is perceptible in the experimental partitions applying subject-level and environmental privacy in \cref{tab:EvaluationAccuracy}. It is worth noting that data from OPERAnet may have reduced environmental heterogeneity as the same hardware, procedure, and subjects were used in a controlled setting. However, this is a common issue for all HAR systems \citep[sec.6]{HARSurvey}. Fused modalities saw an additional drop in accuracy of $2\mbox{-}9\%$, while unfused modalities saw a less significant impact, with the additional maximum drop never exceeding $5\%$. Notably, ResNet34 operating on unfused modalities did not exhibit a significant accuracy drop when privacy was increased. Fused modalities contain more information about the environment per sample, aiding in distinguishing between rooms. However, the additional information becomes less valuable after samples are split into clients. \Cref{fig:CorrImprovLoc} highlights that the small EfficientNetV2 suffers more from not having examples from both rooms available.

\subsection{Modality-level Privacy}

The experimental results reveal an unexpected pattern when applying privacy at the modality level. The previously top-performing ResNet34 experiences a $37\%$ further drop in accuracy with a $42\%$ total, while the EfficientNetV2 suffers a $25\%$ further drop in accuracy with a $32\%$ total. By contrast, the Fusion Vision Transformer (FViT), which produced worse results in previous experiments, only experiences a $14\%$ further drop in accuracy with $19\%$ total and emerges as the model with the most significant performance advantage for a given privacy level. To better understand this outcome and the interplay between different model types, we turn to the plot in \cref{fig:CorrImprovLoc}, which shows the convergence of models for different partitions. The plot immediately reveals the steeper slope of improvement that FViT obtains in the first few rounds. Furthermore, this pattern of performance aligns with the fine-tuning experiments reported in \citep{SelfSupervisedVision}, where FViT outperformed ResNet when both were trained on a small amount ($1\mbox{-}20\%$) of data.

The increasing prevalence of IoT devices, surveillance cameras, personal smartphones, and passive RF sensors has led to extensive human activity recognition (HAR) data collection. However, with no uniform regulation or competitive environment, it is critical to prioritise privacy preservation and address the afferent accuracy degradation.

\subsection{Mutual Learning with Per-modality Group Models}

We evaluate the effectiveness of our ensemble, which uses mutual learning to handle the challenges of federated learning across modalities. As demonstrated in both \cref{fig:CorrImprovLoc} and \cref{tab:EvaluationAccuracy}, the ensemble achieves near-equivalent accuracy to FViT on the "Subj+Env" partition with colocated modalities. However, training the federated learning and group-level models simultaneously is costly and difficult to optimise. Moreover, our hyperparameter search, which explored 79 combinations of distillation weights, revealed that the ensemble's performance is sensitive to hyperparameter changes. Meanwhile, the federated model failed to surpass the "Subj+Env+Mod(S)" result in \cref{tab:EvaluationAccuracy} through mutual learning, primarily due to the inherent difficulty of multimodal training on the same network without employing specific multi-task techniques.
\section{Conclusion}
\label{Conclusion}
We investigated the performance of Multimodal Federated Human Activity Recognition under privacy levels that may arise in practice, such as the subject(user), environmental, and modality levels. Our results show that performance degrades with each additional privacy layer starting with $5\mbox{-}7\%$ for the subject and environmental levels. Remarkably, we observed an \emph{overall} accuracy drop of $32\mbox{-} 42\%$ for CNNs when modality-level privacy is assumed. Nevertheless, our experiments determined that a Fusion Vision Transformer architecture performs well in extreme scenarios. Its fast initial convergence with few samples led to an \emph{additional} drop of only $14\%$ with a $19\%$ \emph{overall} drop for modality-level privacy. Furthermore, constructing small group-level models for each modality type trained in a mutual-learning fashion with a global one can limit the \emph{overall} degradation to $7\mbox{-}13\%$. Such a system can adjust to shifting hardware conditions by incorporating new group-level models and utilising the global model's knowledge for bootstrapping. Despite the clear trends, this work is limited by the size of OPERAnet. Besides larger datasets, other potential future research avenues include hierarchical FL with layered aggregation and creating sparse models with task-based subnetworks.

\bibliography{example_paper}
\bibliographystyle{mlsys2023}

\clearpage
\appendix

\section{Appendix}\label{Appendix:parameters}

\begin{table}[h]
    \centering
    \caption{The F1-Score results of partition-model combinations. The same trends from the accuracy comparisons repeat themselves with higher privacy requirements leading to worse performance. A similar strong decline can be observed when clients are partitioned by modality, with the FViT performing the best in the most heterogeneous condition despite trailing behind the CNNs for all other partitions. The ensemble group models also successfully recovered performance near the FViT levels when the partitioning was based only on subject and environment.}
    \label{tab:Appendix:F1-Score}
    \resizebox{\columnwidth}{!}{%
        \begin{tabular}{@{}lcccc@{}}
            \toprule
            Partition            & FViT                             & ResNet34                          & EffNetB0                         & Ensemble                         \\ \midrule
            Centralised (Fused)  & 0.80$\pm$ 0.03                   & \textbf{0.86} $\pm$ \textbf{0.02} & 0.83$\pm$ 0.02                   & -                                \\
            Subj (Fused)         & 0.73$\pm$ 0.03                   & \textbf{0.82} $\pm$ \textbf{0.05} & 0.78$\pm$ 0.02                   & -                                \\
            Subj+Env (Fused)     & 0.70$\pm$ 0.05                   & \textbf{0.74} $\pm$ \textbf{0.08} & 0.64$\pm$ 0.04                   & -                                \\ \midrule
            Centralised (Split)  & 0.71$\pm$ 0.01                   & \textbf{0.80}$\pm$ \textbf{0.03}  & 0.78$\pm$ 0.02                   & -                                \\
            Subj (Split)         & 0.68$\pm$ 0.01                   & 0.72$\pm$ 0.06                    & \textbf{0.73}$\pm$ \textbf{0.02} & -                                \\
            Subj+Env (Split)     & 0.64$\pm$ 0.03                   & \textbf{0.71}$\pm$ \textbf{0.04}  & 0.67$\pm$ 0.02                   & -                                \\
            Subj+Env+Mod (Split) & \textbf{0.50}$\pm$ \textbf{0.04} & 0.35$\pm$ 0.03                    & 0.39$\pm$ 0.04                   & \textbf{0.60}$\pm$ \textbf{0.03} \\ \bottomrule
        \end{tabular}%
    }
\end{table}

The preprocessing pipeline we use is precisely described in \citet{SelfSupervisedVision}; however, we shall offer a brief summary here. First, the CSI signal is denoised using a discrete wavelet transform and median filtering before applying PCA and generating a spectrogram through the STFT\@. Then, for the PWR data, the authors apply the cross ambiguity function to the PWR data followed by the CLEAN algorithm and the outputting of a Doppler spectrogram. We use three of the image types they generate. First, we use the concatenated spectrograms generated from the three-receiver surveillance channels of the PWR data. The combined images from the three channels have a dimension of $224\times672$. Second, we use the spectrograms generated using STFT on amplitude CSI data from two receivers with a concatenated size of $224 \times 448$. Third, we use the phase-difference spectrograms generated via STFT from the phase-difference CSI data from the two receivers with a concatenated size of $224 \times 448$. Combined in the fused partitions, they add up to $224 \times 1568$ images. Finally, we take the largest image type ($224\times672$) in unfused partitions and pad the rest.

\begin{figure*}[!t]
    \centering
    \includegraphics[clip,width=\textwidth]{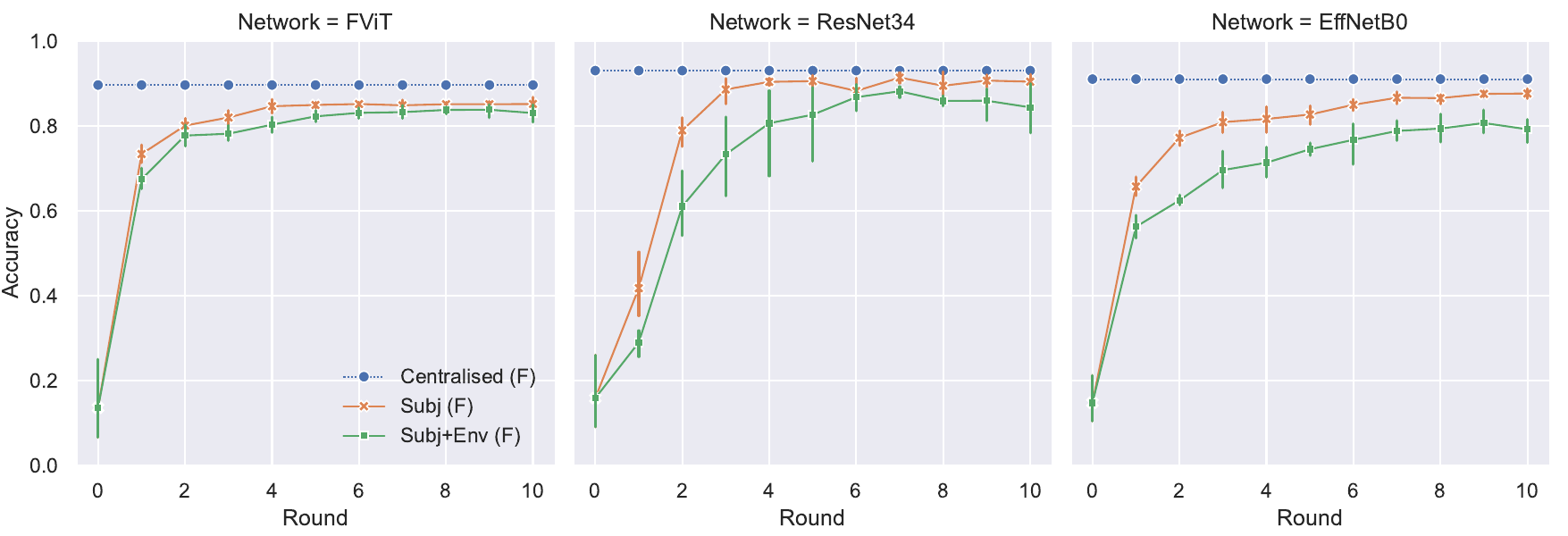}
    \caption[]{Model per-round accuracy on the fused-modality dataset. The trends observed resemble those for the split partitions with only a uniform decrease in accuracy by comparison. The only major change in results is the sensitivity of ResNet34 to environmental privacy.}
    \label{fig:AppendixFigFusedConvergence}
    \vspace{128in}
\end{figure*}

Client data partitions are generated in order of person index for split-subject modalities, person index and then room index for subject and environment, and subject, room, and modality index for the final partitioning. Our indexing assumes the human subjects are ordered from one to six, the rooms from one to two, and the modalities from one to three in the above order. The subject and room indexes are directly available in the dataset. Each model and partition combination was run using five distinct seeds generating the same client sequence across models. Thus differences in performance between models are not due to randomness in client selection. The seeds we use are $42,1337,3407,8711,9370$, and the client sequence is generated by calling {\tt np.random.choice} for the given number of clients per round out of the entire population for each of the ten rounds at the start of the script right after the seed has been set. The mean and standard deviation are reported based on the five seeds in and \cref{tab:EvaluationAccuracy} and \cref{tab:Appendix:F1-Score}. The per-round values in \cref{fig:CorrImprovLoc} and \cref{fig:AppendixFigFusedConvergence} have their mean and standard deviation calculated based on the accuracy of the models on each of the five seeds at the specific round. Before every experiment, the same seeds are used to set the random, NumPy, and Torch modules in Python.

All models have been trained as in \citet{SelfSupervisedVision} using AdamW with $\beta_1=0.9$ and $\beta_2=0.999$ with a weight decay of $0.01$ and batch size of 10 rather than 64 due to the small size of the federated partitions. The computational resources involved four Nvidia A40s and were extensively used during parameter tuning.

\FloatBarrier

\end{document}